\documentclass[conference]{IEEEtran}
\IEEEoverridecommandlockouts
\pdfoutput=1
\usepackage{cite}
\usepackage{amsmath,amssymb,amsfonts}
\usepackage{algorithmic}
\usepackage{array} 
\usepackage{graphicx}
\usepackage{makecell}
\usepackage{algorithm}
\usepackage{tabularx,ragged2e,booktabs}
\usepackage{adjustbox}
\usepackage{algorithmic}
\usepackage{textcomp}
\usepackage[usenames,dvipsnames]{color}
\def\BibTeX{{\rm B\kern-.05em{\sc i\kern-.025em b}\kern-.08em
    T\kern-.1667em\lower.7ex\hbox{E}\kern-.125emX}}
    \newcommand{\ignore}[1]{}
\begin{document}

\title{A Convolutional Spiking Network for Gesture Recognition  in Brain-Computer Interfaces\\
}


\author{\IEEEauthorblockN{Yiming Ai}
\IEEEauthorblockA{\textit{Department of Engineering} \\
\textit{King’s College London}\\
London, United Kindom \\
yiming.ai425@gmail.com}
\and
\IEEEauthorblockN{Bipin Rajendran}
\IEEEauthorblockA{\textit{Department of Engineering} \\
\textit{King’s College London}\\
London, United Kindom \\
bipin.rajendran@kcl.ac.uk}
}

\maketitle

\begin{abstract}
Brain-computer interfaces are being explored for a wide variety of therapeutic applications. Typically, this involves measuring and analyzing continuous-time electrical brain activity via techniques such as electrocorticogram (ECoG) or electroencephalography (EEG) to drive external devices. However, due to the inherent noise and variability in the measurements, the analysis of these signals is challenging and requires offline processing with significant computational resources. In this paper, we propose a simple yet efficient machine learning-based approach for the exemplary problem of hand gesture classification based on brain signals. We use a hybrid machine learning approach that uses a convolutional spiking neural network employing a bio-inspired event-driven synaptic plasticity rule for unsupervised feature learning of the measured analog signals encoded in the spike domain. We demonstrate that this approach generalizes to different subjects with both EEG and ECoG data and achieves superior accuracy in the range of 92.74-97.07\% in identifying different hand gesture classes and motor imagery tasks.


\end{abstract}



\section{Introduction}
\textbf{Context and motivation.} Brain-computer interface (BCI) is a  cutting-edge technology that adopts state-of-the-art signal processing methods and machine learning algorithms to detect underlying patterns in time-varying signals measured from the brain. 
Nerve illnesses due to neurological disorders, nerve pathway blockage, poliomyelitis or strokes, and muscular impairments like paralysis, amyotrophy, or  accidents all cause both physical and mental injuries to patients. In this scenario, BCI provides a potential future for these patients to alleviate or even eliminate the disability and improve their quality of life by enabling them to use motor imagery to operate computers, robotic prostheses, and even their own organs\cite{leuthardt2004brain,citi2008p300,kim2008neural,curran2003learning,mcfarland2008brain}. One of the most crucial components of a reliable BCI technology is the interpretation of measured biological signals  to drive prosthetic devices. 
How to best correlate motor imagery -- the cognitive process involving the  imagination of a movement without actually executing it, and the actual motor intention -- remains a challenging problem \cite{liang2012decoding,alazrai2019eeg,kubanek2009decoding}.

\begin{figure}[!h]
\centerline{\includegraphics[width=\linewidth]{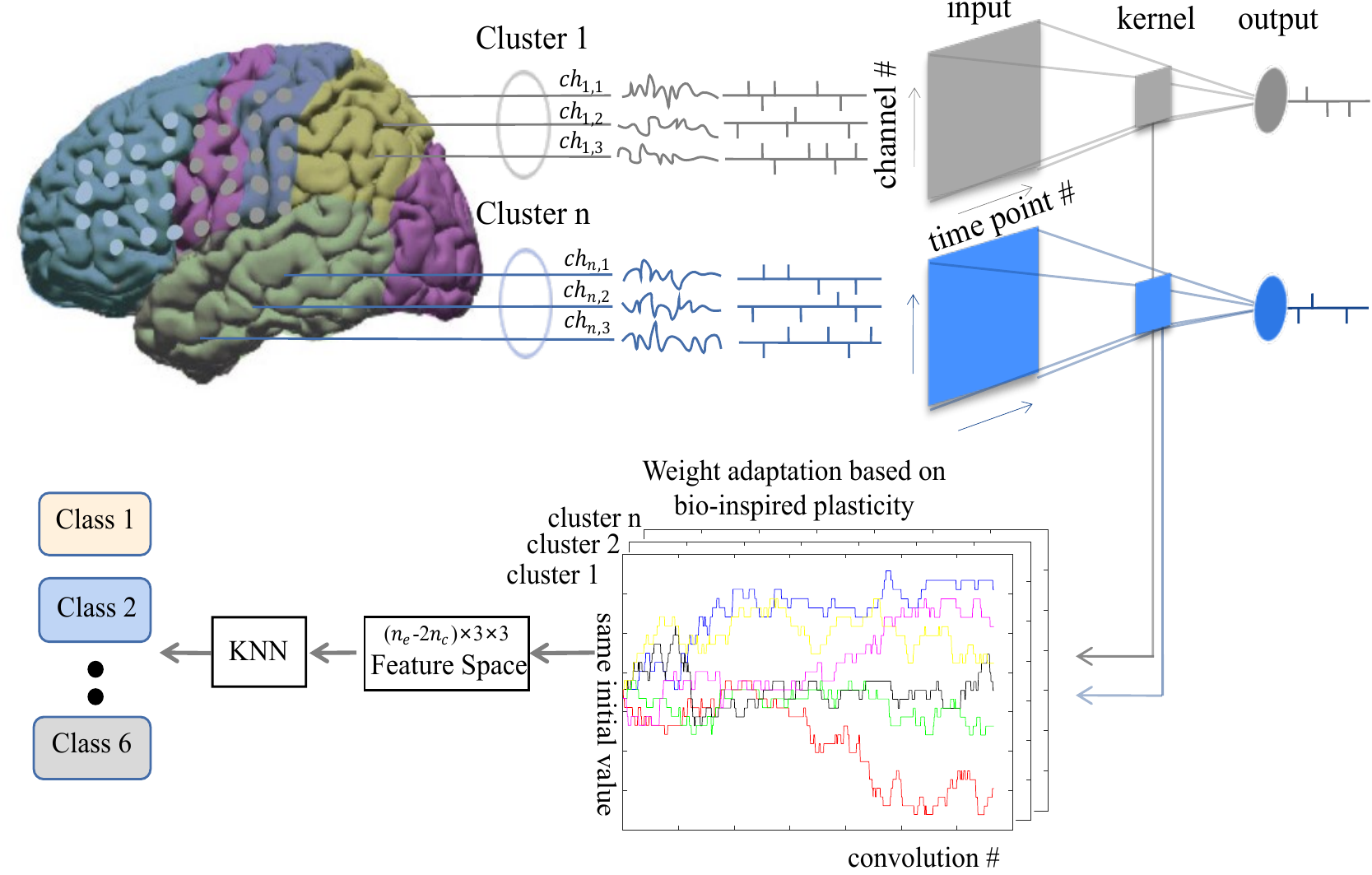}}
\caption{The proposed overall scheme for spatio-temporal signal analysis from BCI recordings.}
\label{fig:flowchart}
\end{figure}

Traditional signal processing methods as well as machine learning algorithms \cite{zhang2019survey,gao2020deep} have made remarkable progress in solving BCI-based behavior recognition tasks \cite{tangermann2012review}, although efficient analysis of neural signals that are  time-varying and multi-dimensional remains a challenging task.  Combining the abilities  of convolutional neural  networks  (CNN) for pattern recognition and recurrent neural networks (RNN) for sequential signal processing, a hybrid CNN-RNN architecture has been implemented for surface electromyography (sEMG)-based gesture recognition \cite{hu2018novel}. As the size of the measured data  increases, aforementioned deep learning algorithms that use a large parameter space make real-time predictions with high accuracy and reliability across different subjects particularly challenging.



Spiking neural networks (SNNs) are machine learning models that mimic the spike-triggered information processing modality of the  brain \cite{maass1997networks}. They have some natural advantages in the processing of time-varying signals as spiking neuron models naturally incorporate recurrent dynamics. They also hold promise for event-driven learning and adaptation as well as continual learning \cite{rosenfeld2022spiking}\cite{chen2023neuromorphic}. They are also widely studied for efficient hardware implementations using both CMOS and post-CMOS device technologies \cite{rajendran2019low}.   

\textbf{Main contributions.} 
In this paper, we develop a hybrid machine-learning approach with high classification accuracy that could be used for real-time gesture recognition. Thanks to the small number of parameters involved in the  model, this method is also amenable to implementation in  battery-operated  embedded processors.


Our approach  treats  neural signals  belonging to different cortical regions separately while extracting temporal correlation information inherent to different time scales. 
We highlight some key features of our approach:

\begin{itemize}
\item  We first  identify electrode channels that belong to the same neighborhood in the 3D-euclidean space and consider the measured signals from the cluster jointly for further processing. 
\item  We apply a temporal contrast coding scheme to translate the measured analog signals to spike streams that consist of positive and negative spikes.
\item  Signals from the same cluster are then analyzed using a spatiotemporal convolution filter whose output drives an integrate-and-fire neuron.
\item  The kernel weights of the convolution filter are modified in an event-driven manner inspired by the biological spike-timing-dependent-plasticity (STDP) rule. 
\item The weights at the end of the presentation of a signal from all the clusters are then analyzed using a k-nearest neighbor (KNN) classifier to identify the gesture class. 
\end{itemize}

A key advantage of our scheme is that the kernel weights of the network evolve in an event-driven manner based on the history of spikes over a very
short duration (3 most recent time instants), thus avoiding the need to record and store analog signals for long durations  for signal processing.

This approach is used to classify gesture signals from the Stanford ECOG dataset and GigaScience EEG data set. For the subjects with electrode placement information, we obtain classification accuracy in the ranges of  $92.74-97.07\%$, surpassing the performance of previously reported work in the literature   \cite{kumarasinghe2021brain,zhang2019novel,zhang2023multi,olivas2019classification,kumar2021optical+,stieger2021benefits,yao2022fast,lee2022individual,zhu2020resot}. 


\section{Background}

 The ECoG signals  used in this work are acquired from the public Stanford dataset library\cite{miller2012human}; we limit our study to the two subjects from the dataset for whom the electrode placement information was available.  To assess brain activity, an array of platinum electrodes were placed on various cortical regions. The subjects were instructed to move a single finger in response to instructions displayed on a computer next to them. For instance, when the computer displayed ``thumb'', the patients were asked to move the corresponding finger. The ECoG signals were gathered over a period of time at a sampling rate of $1000\,$Hz. Finger flexion was recorded during the $2-$second movement trials, where the computer  displays randomly-chosen instructions $2$  seconds per time and the patients  performed self-paced movements $2-5$ times throughout the trial. There was a $2-$second rest interval between two instructions, during which the computer displayed complete darkness. For each finger on each patient, the entire dataset captured finger flexion 30 times.

Additionally, five EEG subjects are selected from the GigaScience database \cite{cho2017eeg}, where the motor imagery-based neural signals are collected in a similar manner. 

\begin{figure}[h]
\centerline{\includegraphics[width=\linewidth]{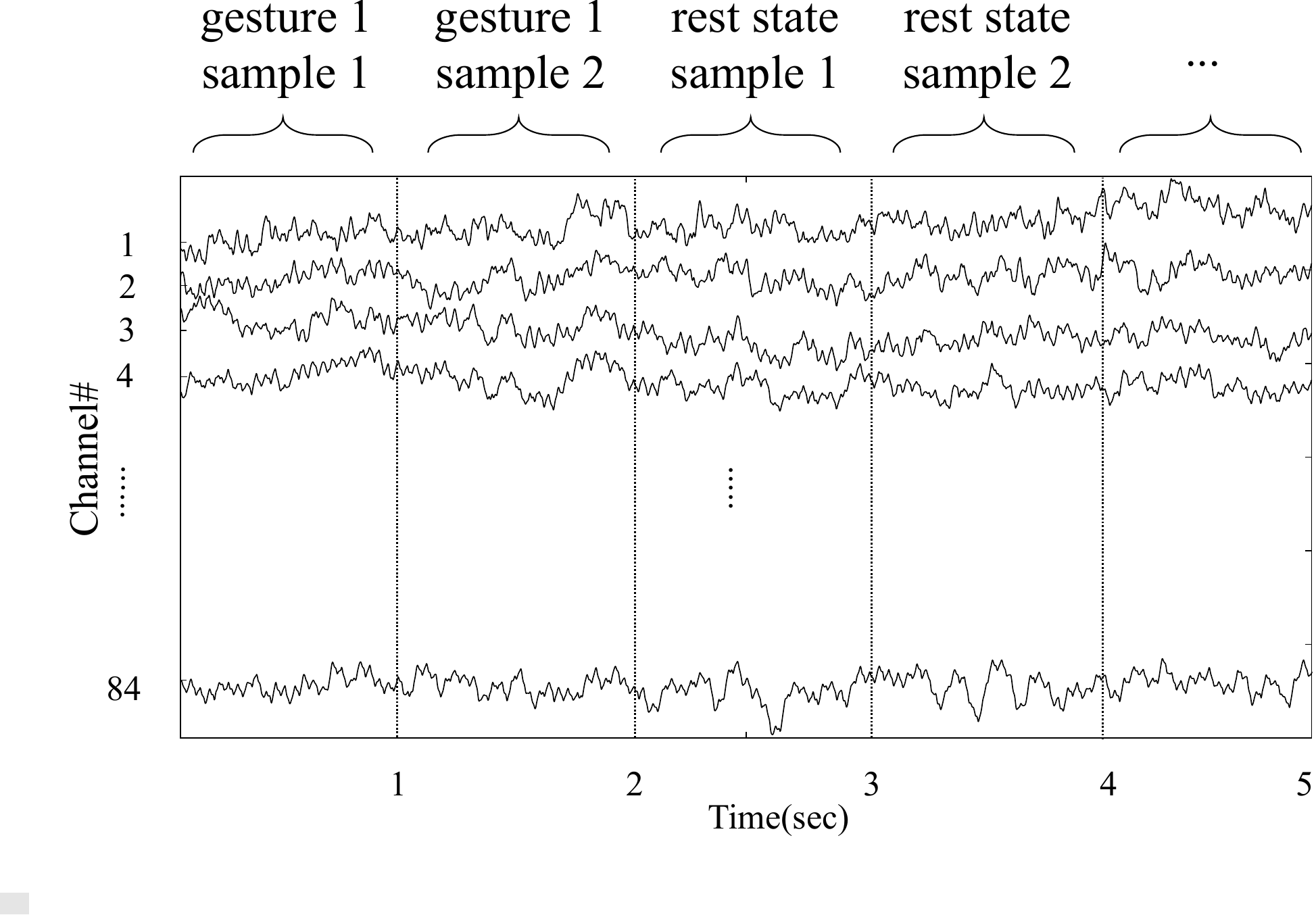}}
\caption{Each sample for training our model is obtained from one second of the recording of the neural signals.}
\label{fig:experiment}
\end{figure}

To train our models, we split the neural signals from the whole trial into multiple segments. While each gesture recording lasted $2$ seconds, we have split the signals in each two-second duration into two samples corresponding to the same gesture (Figure \ref{fig:experiment}). In total, we obtain $400$ samples for each subject.

	
	

\section{Methods}
We now describe the different steps in our approach for gesture classification. 

\textbf{Spike encoding.} We first encode the continuous time analog brain signal to a discrete spike train with positive and negative spikes as per the standard temporal contrast (TC) coding scheme \cite{bohte2002unsupervised}. As will be discussed later, we use the sign of the spike to trigger positive and negative weight adaptation during the learning phase.

The TD encoding algorithm is implemented as follows. Given a fixed threshold $\theta_{th}$, for every time point starting from zero, the difference between the value of the analog signal at the  current time step $t_k$ and the previous time step $t_{k-1}$ is calculated as $\Delta u_k=f(t_k)-f(t_{k-1})+u_{k-1}$. The absolute value of $\Delta u_k$ is then compared with a threshold $\theta_{th}$. If $\Delta u$ is larger than the $\theta_{th}$, a spike is issued at the current time instant, and the sign of the spike is set to be the same as the sign of $\Delta u_k$. If $\Delta u_k$ is less than the $\theta_{th}$, no spike will be issued.  Also, when a spike is issued,  $\Delta u_k$ is reset to 0. 





\begin{figure}[!h]
\centerline{\includegraphics[width=\linewidth]{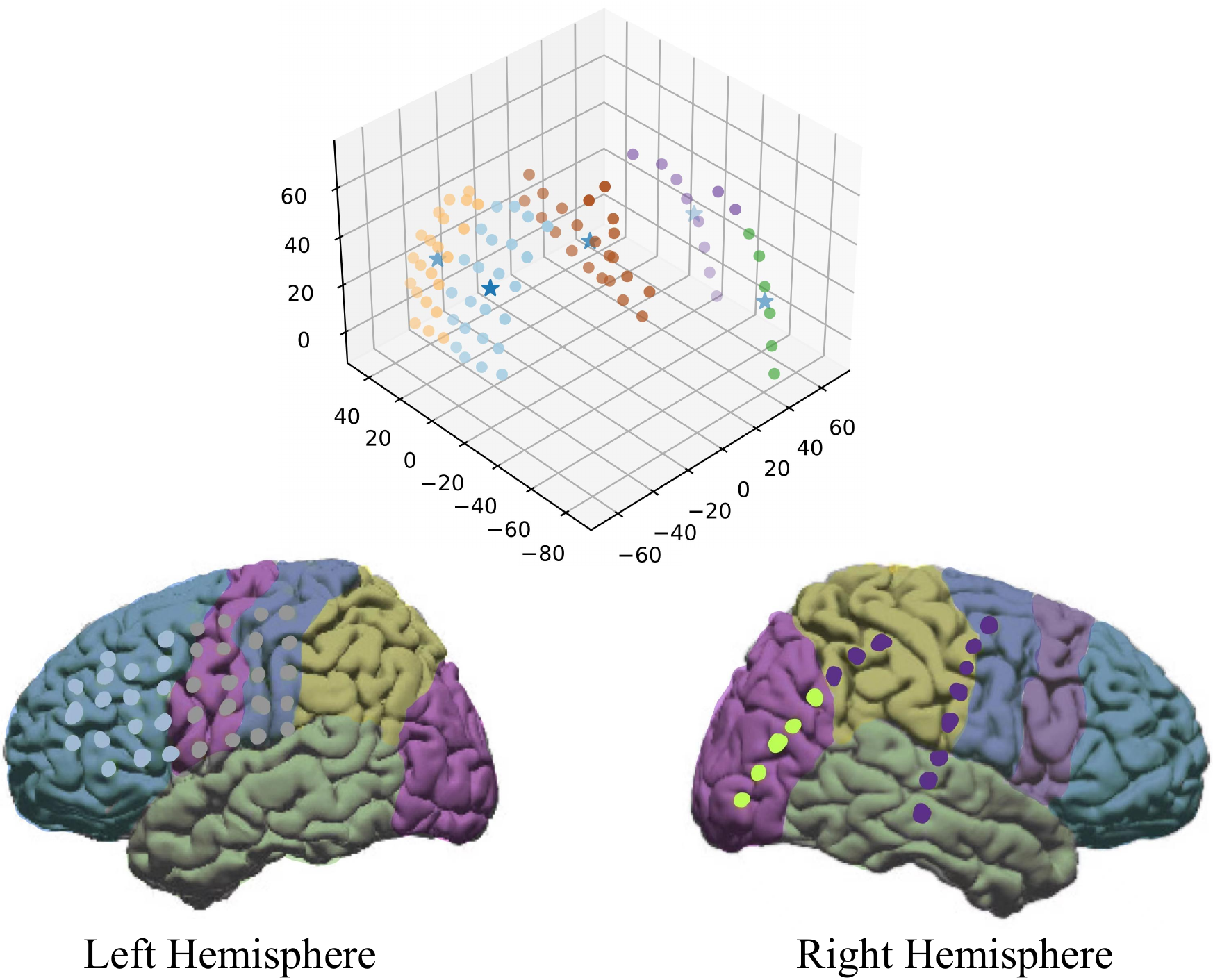}}
\caption{Spatial cluster outcome as well as its 3D cortical projection according to the `location' variable for one of the subjects from the ECoG dataset.}
\label{fig:cluster}
\end{figure}

\textbf{Spatial clustering of electrode channels.} We next employ a spatial clustering approach as a means  to determine neuronal recordings that may be closely correlated and to avoid combining measurements that are from vastly separated regions. For the ECoG signals, the standard k-means clustering algorithm is used to partition the electrodes into subgroups based on their location in the euclidean space. Figure \ref{fig:cluster} illustrates $n_c=5$ clusters obtained for one of the subjects in the ECoG dataset using the standard Elbow method.

\begin{figure}[h!]

\centering
\includegraphics[width=.8\linewidth]{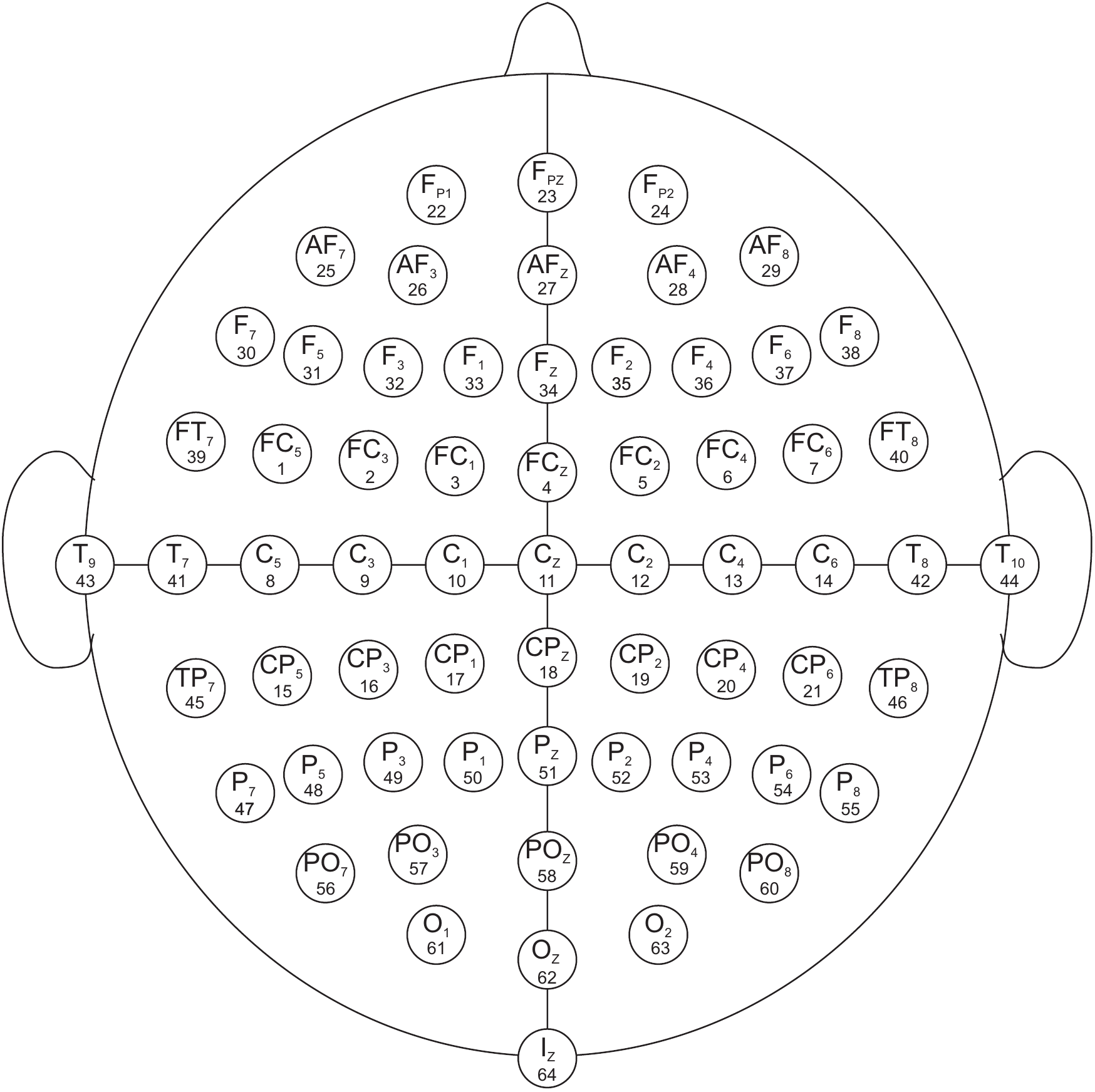}
\caption{\label{fig:BCI2000}EEG channel configuration, adapted from \cite{qiu2020brain}. }
\end{figure}

For the data from the EEG dataset, the spatial clustering step using k-means algorithm can be omitted as the clusters can be obtained automatically using the spatial distribution information available for the 64 channels (Fig. \ref{fig:BCI2000}).




\begin{figure}[!h]
\centerline{\includegraphics[width=\linewidth ]{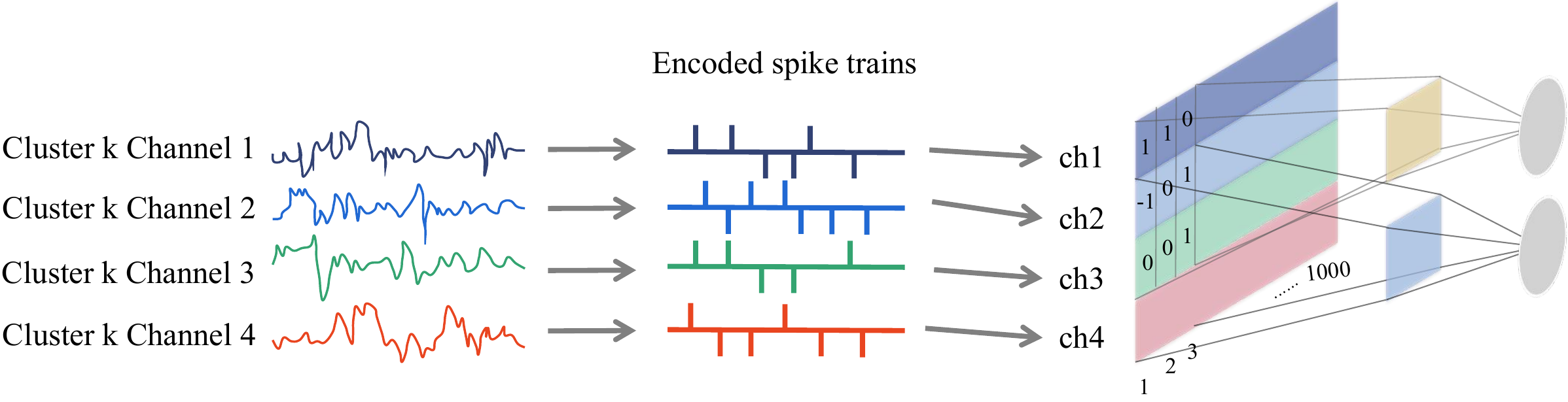}}
\caption{Illustration of the convolutional SNN implementation.}
\label{fig:convolution}
\end{figure}

\textbf{Convolutional spiking neural network.} The spike signals from all the channels in each cluster are then processed by a convolutional spiking neural network. We illustrate our approach in Fig. \ref{fig:convolution} for an exemplary case of 4 channels in a cluster.  The spike values from the channels   \{-1,0,1\} are unfolded along the $1000$ timesteps of the temporal dimension for each sample and  convolved with a $3\times3$ convolution kernel. We use a distinct kernel with the same initial random values for every three channels in the spatial dimension with a stride of 1, which means that each kernel will only stride across the temporal dimension once and record the spike dynamics in three adjacent channels. The output of the kernel feeds into a standard integrate-and-fire spiking neuron.




\textbf{Bio-inspired unsupervised learning.} In order to distinguish the spiking dynamics from different samples, a modified event-driven weight adaptation learning rule is employed  to update the convolution kernel. The  learning  in our  model is triggered when  the postsynaptic neuron spikes as a result of the integration of the spikes through the convolutional filter during the feedforward computation. The detailed description of the algorithm is as follows. 


At every convolution step $t$, if the output neuron issues a spike, we update the  weights corresponding to all the pixels of the kernel that have an incident spike.  The kernel weight is potentiated if the output spike is positive according to the following rule depending on the  sign of the input spike:
\begin{equation}
 \Delta W=   \begin{cases}
\exp(\tau-t-\tau_r) &\text{if input spike is positive,}\\
\exp(-(\tau-t-\tau_r)^2) &\text{if input spike is negative.}\\
\end{cases}
\end{equation}

Here, $\tau$ is the temporal location of the input spike relative to the current time instant, $t-\tau=0, 1, 2$ as we use a $3\times3$ kernel;  $\tau_r$  represents a fixed learning rate which is a hyperparameter for our model. If the output spike is negative, the weights are depressed by the same magnitude as above. If there are no spikes at the output neuron, the kernel weights remain unchanged. The membrane potential of the output neuron does not have a leak and is reset only when there is a spike.

\textbf{KNN Classifier.} After the input spikes for the full duration are presented and the kernel weights updated in an event-triggered manner as described above, each sample is represented by a concatenated kernel weight vector of dimension $[3\times3\times(n_e-2n_c)]\times1$. In the classification phase, these kernel features are fed into a standard KNN classifier along with the labels of the gesture.  During the testing phase, the kernel feature vectors are similarly   calculated using the above-described weight update rule, which is then fed to the KNN classifier to obtain a decision class. 



\section{Results}

\begin{figure}[h]
 \centering \includegraphics[width=\linewidth]{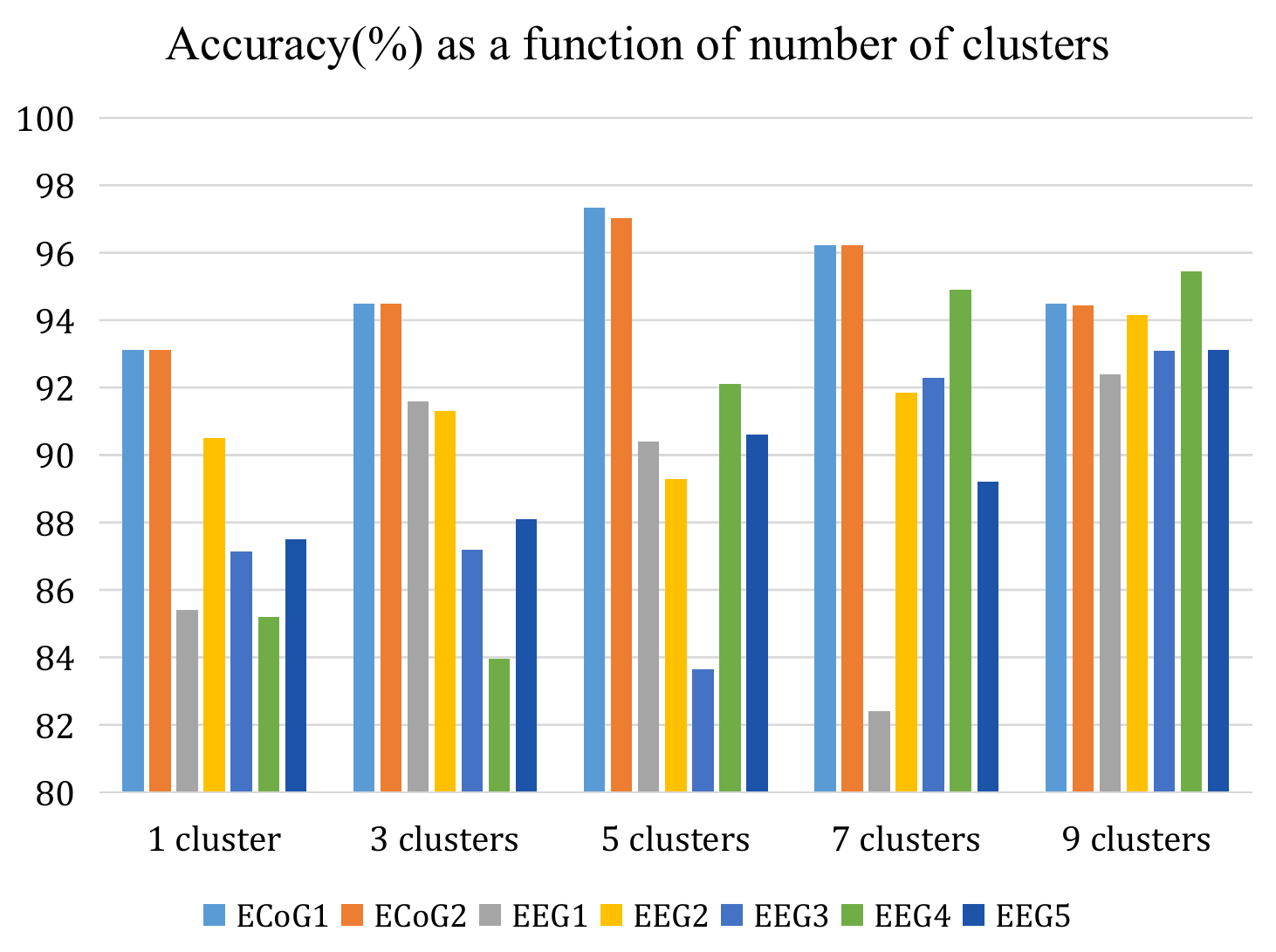}
\caption{\label{fig:clustereffect} Classification accuracy as a function of the number of clusters to which the electrodes are grouped. The best performance is obtained with 5 clusters for the ECOG data and 9 clusters for the EEG data.}
\end{figure}

Datasets obtained from the Standford Human Cortical Activity \cite{miller2012human} as well as GigaScience database Library \cite{cho2017eeg} are used for evaluating the proposed algorithmic framework. We first report the effect of the number of clusters on the classification accuracy (Fig. \ref{fig:clustereffect}). In our study, the best classification accuracies were obtained when the two subjects in the ECOG dataset were partitioned into $n_c=5$ clusters and the five subjects from the EEG dataset into $n_c=9$ clusters. The number of electrodes or channels varies between $6$ and $24$ in the ECOG subjects, and between $3$ and $9$ in EEG subjects.  

\begin{figure}[h]
 \centering \includegraphics[width=\linewidth]{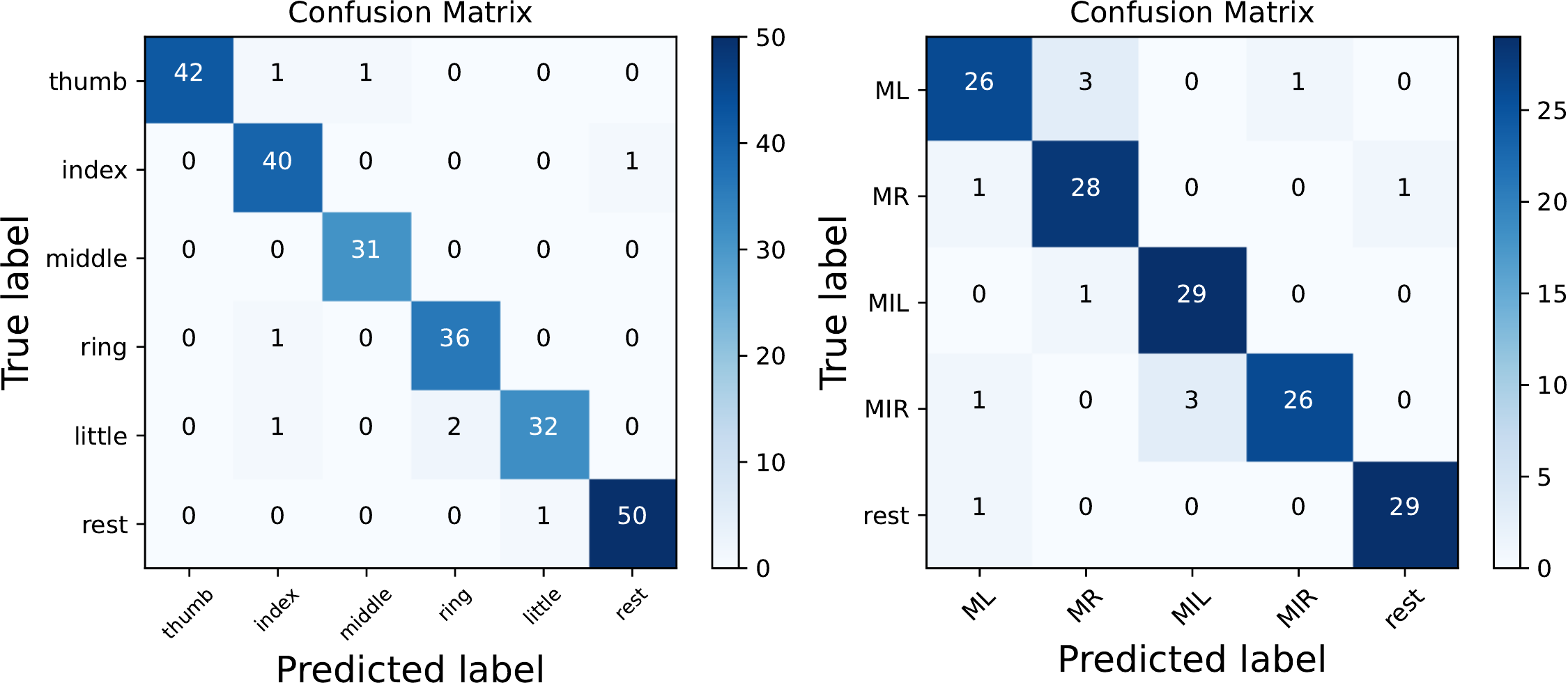}
\caption{\label{fig: confusion}Combined  confusion matrix for the two subjects from the ECoG dataset (left) and 5 subjects from the EEG dataset (right). `ML' is left hand movement, `MR' is right hand movement, `MIL' is motor imagery left hand, `MIR' is motor imagery right hand.}
\end{figure}

For the two subjects selected from the ECoG dataset, we report a classification accuracy of $97.07\%$ (subject `bp'), and
$95.85\%$ (subject `ca'). For the EEG data, with minor adjustments to the hyperparameters, our methodology results in  classification accuracy  ranging between $92.74\%$ to $96.51\%$  for $5$ different subjects in this dataset. The confusion matrix for the 6 gestures is reported in Fig. \ref{fig: confusion}. The values of hyperparameters used in our simulations are presented in table \ref{tab:parameter}.



To benchmark our results, we report  results from the literature that has applied various signal processing techniques for similar BCI classification tasks (Table \ref{tab:comparison}). To the best of our knowledge, the algorithmic framework we have presented achieves superior accuracy compared to all the prior art. 

\begin{table}[h]
\renewcommand\arraystretch{1}
\tiny
\caption{\label{tab:parameter}List of Hyperparameters used in our simulation.} \vspace{- 1.5em}
\begin{center}
\begin{adjustbox}{width=\columnwidth}
\begin{tabular}{ccc}
\toprule
	Parameter & ECoG & EEG\\
\midrule
    Encoding Threshold ($\theta_{th}$)& 0.22 & 0.18\\
    K-means Cluster Number ($n_c$) & 5 & 9\\
    Kernel Dimension & 3$\times$3 & 3$\times$3 \\
    Temporal Dimension Stride & 3 & 3\\
    Channel Dimension Stride & 1 & 1\\
    Unsupervised Learning Rate ($\tau_r$)& 5 & 6\\
    Convolution Threshold & 0.1 & 0.12\\
    KNN's K Value & 5 & 5\\
\bottomrule

\end{tabular}
\end{adjustbox}
\end{center}
\end{table}

\vspace{ .1in}

\begin{table}[h]
\renewcommand\arraystretch{3.5}
  \vspace{-1.5em}
\caption{\label{tab:comparison}Comparison of  our work with state-of-art BCI classification.} \vspace{- 1.5em}
\begin{center}
\begin{adjustbox}{width=\columnwidth}
\begin{tabular}{ccccc}

\toprule[2pt]
	Reference & Task & Probe Location & Method & {\makecell[c]{Accuracy (\%) \\ or Cross-correlation(cc)}} \\
\midrule

    \cite{kumarasinghe2021brain}& Hand movement & EEG & BI-SNN & 0.55-0.74 (cc)\\

    \cite{zhang2019novel} & Hand movement  & EEG & CNN and WNN  & 77.9\%-90.1\% \\
 
    \cite{zhang2023multi} &Hand movement   & EEG  & MSGMTL & 73.10\% \\
    
    \cite{olivas2019classification} & Hand movement   & EEG  & expert CNN & 73.78\%-91.17\% \\

    \cite{kumar2021optical+} & Hand movement   & EEG  & OPTICAL+ & 69.59\% \\

    \cite{stieger2021benefits} & Hand movement   & EEG  & CNNs & 68.19\% \\
    
    \cite{yao2022fast} & Finger Flexion   & {\makecell[c]{Primary Motor \\ Cortex ECoG}}  & Temporal lightGBM & 77.0\% \\

    \cite{lee2022individual} & Finger movement  & {\makecell[c]{Primary Motor \\ Cortex ECoG}} & linear SVM & 60.6\%-70.7\% \\

    \cite{zhu2020resot} & Finger movement  & {\makecell[c]{Primary Motor \\ Cortex ECoG}}& ResOT-$L_2$ & 61.3\% \\

	Our Work & Finger Flexion & {\makecell[c]{Primary Motor \\ Cortex ECoG}} & {\makecell[c]{Convolutional SNN- \\ based Spatiotemporal \\ Signal Analysis }}& {\makecell[c]{97.07\% and \\ 95.85\% for \\ two subjects }} \\

	Our Work & {\makecell[c]{Motor Imagery\\  Hand Movement}} & EEG & {\makecell[c]{Convolutional SNN- \\ based Spatiotemporal \\ Signal Analysis }}& {\makecell[c]{92.74\%-96.51\% \\ for five subjects}} \\
\bottomrule[2pt]
 \vspace{-6em}
\end{tabular}
\end{adjustbox}
\end{center}
\end{table}

\section{Conclusion}


Time-varying signals obtained from BCI techniques such as ECOG and EEG encode  rich spatiotemporal patterns capturing motor imagery, though their analysis is challenging due to subject-to-subject variations and inherent noise in measurements. In this paper, we  proposed an efficient algorithmic framework for the classification of finger gestures based on two exemplary datasets, that combine spatial clustering of the measurement channels and temporal analysis of the spike-encoded signals based on a convolutional spiking neural network. The kernel weights of the network evolve in an event-driven manner based on the history of spikes over a short duration, which can then be used to identify the gesture class with superior accuracy compared to methods in the  prior art. Hence, our method can be used for \textit{in situ} and real-time classification of motor imagery based on locally available spatial and temporal information for BCI applications.

\section*{Acknowledgment}

  This research was supported in part by an Open Fellowship of the EPSRC (EP/X011356/1).
\newpage

\bibliographystyle{unsrt}
\bibliography{ref}

\end{document}